\documentclass{article} 
\usepackage{iclr2025_conference,times}

\usepackage{amsmath,amsfonts,bm}

\def\eqref#1{equation~\ref{#1}}

\def\1{\bm{1}}

\def\va{{\bm{a}}}

\def\vu{{\bm{u}}}
\def\vv{{\bm{v}}}

\def\vx{{\bm{x}}}

\def\vz{{\bm{z}}}

\DeclareMathAlphabet{\mathsfit}{\encodingdefault}{\sfdefault}{m}{sl}
\SetMathAlphabet{\mathsfit}{bold}{\encodingdefault}{\sfdefault}{bx}{n}

\def\sA{{\mathbb{A}}}

\def\sR{{\mathbb{R}}}
\def\sS{{\mathbb{S}}}

\usepackage{hyperref}
\usepackage{url}
\usepackage{makecell}
\usepackage{soul}

\usepackage{booktabs}
\usepackage{multirow}
\usepackage{graphicx}
\usepackage{wrapfig}
\usepackage{subcaption}
\usepackage{adjustbox}
\usepackage{algpseudocode,algorithm,algorithmicx}

\title{Model-Agnostic Knowledge Guided Correction for Improved Neural Surrogate Rollout}

\author{Bharat Srikishan\textsuperscript{1}\thanks{corresponding author} , Daniel O'Malley\textsuperscript{2}, Mohamed Mehana\textsuperscript{2}, Nicholas Lubbers\textsuperscript{2}, \\ \textbf{Nikhil Muralidhar\textsuperscript{1}} \\
\textsuperscript{1}Stevens Institute of Technology,
\textsuperscript{2}Los Alamos National Laboratory \\
\texttt{\{bsrikish,nmurali1\}@stevens.edu, \{omalled,mzm,nlubbers\}@lanl.gov} \\
}

\newcommand{\ourmethod}{Hybrid PDE Predictor with RL}
\newcommand{\ourmethodshort}{HyPER}
\newcommand{\unet}{UNet}
\newcommand{\fno}{FNO}
\newcommand{\unetonly}{UNet-P}
\newcommand{\fnoonly}{FNO-P}
\newcommand{\unetmultistep}{UNet-Multistep}
\newcommand{\mpp}{MPP}
\newcommand{\mppzero}{MPP-ZS}
\newcommand{\refiner}{PDE-Refiner}
\newcommand{\randompol}{Random Policy}
\newcommand{\surrogateonlyshort}{SUG}

\newcommand{\hide}[1]{}
\newcommand{\pluseq}{\mathrel{{+}{=}}}

\iclrfinalcopy 
\begin{document}

\maketitle

\begin{abstract}
Modeling the evolution of physical systems is critical to many applications in science and engineering. As the evolution of these systems is governed by partial differential equations (PDEs), there are a number of computational simulations which resolve these systems with high accuracy. However, as these simulations incur high computational costs, they are infeasible to be employed for large-scale analysis. A popular alternative to simulators are neural network \emph{surrogates} which are trained in a data-driven manner and are much more computationally efficient. However, these surrogate models suffer from high rollout error when used autoregressively, especially when confronted with training data paucity. Existing work proposes to improve surrogate rollout error by either including physical loss terms directly in the optimization of the model or incorporating computational simulators as `differentiable layers' in the neural network. Both of these approaches have their challenges, with physical loss functions suffering from slow convergence for stiff PDEs and simulator layers requiring gradients which are not always available, especially in legacy simulators. We propose the Hybrid PDE Predictor with Reinforcement Learning (\ourmethodshort{}) model: a model-agnostic, RL based, cost-aware model which combines a neural surrogate, RL decision model, and a physics simulator (with or without gradients) to reduce surrogate rollout error significantly. In addition to reducing in-distribution rollout error by \textbf{47\%-78\%}, \ourmethodshort{} learns an intelligent policy that is adaptable to changing physical conditions and resistant to noise corruption. Code available at \url{https://github.com/scailab/HyPER}.
\end{abstract}

\section{Introduction}
\label{sec:intro}

Scientific simulations have been the workhorse enabling novel discoveries across many scientific disciplines. However, executing fine-grained simulations of a scientific process of interest is a costly undertaking requiring large computational resources and long execution times. In the past decade, the advent of low-cost, efficient GPU architectures has enabled the re-emergence of a powerful function approximation paradigm called deep learning (DL). These powerful DL models, with the ability to represent highly non-linear functions can be leveraged as \emph{surrogates} to costly scientific simulations. Recently, the rapid progress of DL has greatly impacted scientific machine learning (SciML) with the development of neural surrogates in numerous application domains. Some highly-impactful applications include  protein structure prediction, molecular discovery~\cite{schauperl2022ai,smith2018transforming} and domains governed by partial differential equations (PDE)~\cite{brunton2024promising,raissi2019physics,lu2021deepxde}. Neural surrogates have also been successfully employed for modeling fluid dynamics in laminar regimes like modeling blood flow in cardiovascular systems~\cite{kissas2020machine} and for modeling turbulent~\cite{duraisamy2019turbulence} and multi-phase flows~\cite{muralidhar2021phyflow,raj2023comparison,siddani2021machineflow}. 

\textbf{Neural Surrogates are Data Hungry}. Although neural surrogates are effective at modeling complex functions, this ability is usually conditioned upon learning from a large trove of representative data. The data-hungry nature of popular neural surrogates like neural operators is well known in existing work \citep{li2020fno,lu2021learningdeeponet, tripura2024multidatahungry,howard2023multifidelitydatahungry,lu2022multifidelitydatahungry}. However, many scientific applications suffer from \emph{data paucity} due to the high cost of the data collection process (i.e., primarily due to high cost of scientific simulations). Hence, neural surrogates employed to model a scientific process of interest, need to address the data paucity bottleneck by learning effectively with a low volume of training data. 

\textbf{Rollout Errors in Neural Surrogates.} Although computational simulations have been designed for modeling various types of physical systems, those exhibiting transient dynamics are especially challenging to model. Solutions to systems exhibiting transient dynamics are usually 
\begin{wrapfigure}{r}{0.47\textwidth}
\vspace{-2ex}
  \begin{center}
    \includegraphics[width=0.42\textwidth]{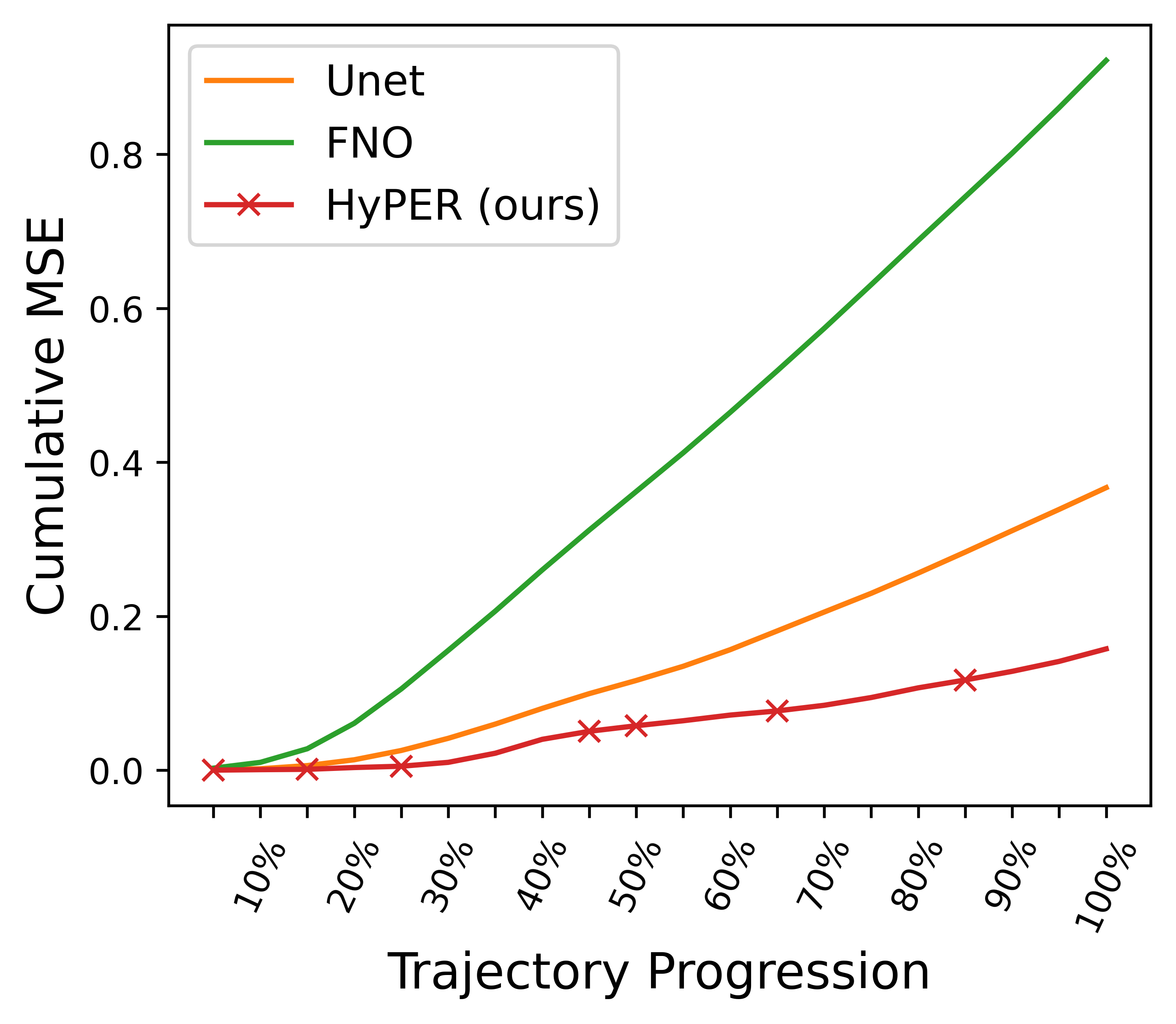}
  \end{center}
  \vspace{-2ex}
  \caption{Cumulative MSE depicting \emph{rollout error} for a single trajectory of \ourmethodshort{} vs surrogate only methods. x's mark the timesteps during the trajectory where our RL policy calls the simulator.\vspace{-2ex}}
\label{fig:ns-cum-mse-surr-rl}
\end{wrapfigure}
 obtained by discrete-time evolution of the dynamics. Simulators used to model such systems are invoked autoregressively and thereby encounter numerical instability and error buildup over long estimation horizons. Such error buildup during autoregressive invocation is termed \emph{rollout error}. Effective techniques have been developed to reduce rollout error of computational simulations and increase their numerical stability over long rollouts. Although autoregressive rollout of neural surrogates is also affected by rollout error, solutions to minimize this error buildup have not been widely investigated. Recently,~\citep{margazoglou2023stability} has inspected the stability of echo-state networks during autoregressive rollout and~\cite{list2024temporal,carey2024data,lippe2024pde} have characterized rollout errors in more general neural surrogates. However, a systematic solution to alleviate rollout error in neural surrogates for modeling transient dynamics is still lacking.

\textbf{Knowledge-Guided Neural Surrogates.} One popular method of addressing errors due to data paucity in neural surrogates is to leverage knowledge of the theoretical model governing the scientific process.
Previous efforts have incorporated domain knowledge (as ODEs, PDEs) while training the DL surrogate to develop knowledge-guided learning pipelines~\citep{raissi2019physics,karpatne2022knowledge,rackauckas2020universal,gao2021phygeonet}. Most of these approaches incorporate the PDE governing the system dynamics as soft regularizers while training the neural surrogates. A majority of such approaches exhibit slow convergence and catastrophic failures in challenging, stiff PDE conditions~\citep{krishnapriyan2021characterizing,wang2022pinnfailure}. 

\textbf{Hybrid-Modeling} All approaches discussed thus far are so-called \emph{surrogate-only} (\surrogateonlyshort{}) approaches. Here, the computational simulator is employed only as a means of generating data to train the neural surrogate and discarded post the training. \surrogateonlyshort{} approaches employ only the pre-trained neural surrogate during inference. Although~\surrogateonlyshort{} provide instantaneous responses relative to computational simulations, they generally have limited generalization ability outside the domain of the training data. An effective complement to~\surrogateonlyshort{} approaches are \emph{hybrid-modeling} approaches~\citep{kurz2022hybrid,karpatne2022knowledge}, that jointly resolve a query by incorporating surrogates in conjunction with computational solvers. Otherwise stated, hybrid-modeling pipelines employ a `solver-in-the-loop'~\citep{um2020solver} approach.
In addition to neural surrogates, there exist a number of classical hybrid modeling techniques which combine a full order model (FOM) with a reduced order model (ROM) such as proper orthogonal decomposition \citep{willcox2002balanced}, dynamic model decomposition \citep{kutz2016dynamic}, and multi-scale methods. While these methods are used to accelerate scientific simulations, they are often limited in expressivity, especially in modeling complex non-linear dynamics. Recent hybrid models \citep{suh2023accelerating} generally have a static coupling between the components in the model and require a static interaction/transition between the FOM and the ROM, while our method proposes an adaptable and learnable interaction between the neural surrogate and the simulator. Our proposed method allows dynamic integration of an FOM (fine-grained simulator) with ROM (neural surrogate) using reinforcement learning.

\textbf{Knowledge-Guidance with Hybrid-Modeling.} Hybrid-modeling approaches are inherently knowledge-guided. A majority of the recent hybrid-modeling approaches are based on directly incorporating PDE solvers as additional layers in the deep learning architecture of neural surrogates~\citep{chen2018neural,belbute2020combining,donti2021dc3,pachalieva2022physics}. While such approaches address the issues of large errors under data paucity and during rollout, inherent in~\surrogateonlyshort{}  approaches, they impose the crippling restriction of \emph{differentiability} on the computational solvers to be incorporated as part of the DL pipeline. Most solvers and computational simulations are NOT differentiable out-of-the-box and hence imposing such differentiability constraints drastically curtails the applicability of current hybrid-modeling approaches.

\begin{figure}[!t]
\begin{center}
\includegraphics[width=0.95\textwidth]{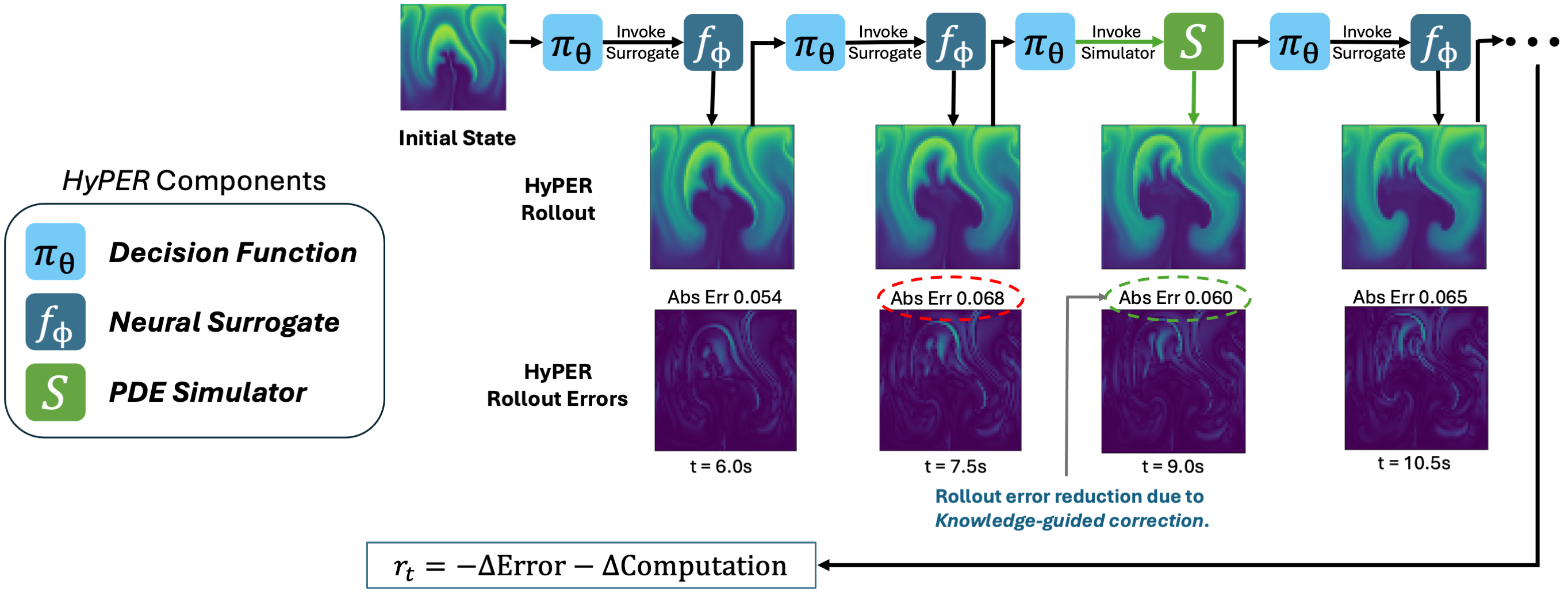}
\end{center}
\caption{Overview of \ourmethod{} (\ourmethodshort) with example rollout. Here $\pi_{\theta}$ is the decision model, $f_{\phi}$ is the surrogate, and $S$ is the simulator. At $t = 9.0$ in the above trajectory, the decision policy invokes the simulator, correcting the trajectory to reduce rollout error. The effect of this knowledge-guided correction can be observed by a reduction in absolute error (dotted green circle) in the figure.\vspace{-3ex}}
\label{fig:hyper-overview}
\end{figure}

To address the existing challenges with surrogate-only and hybrid modeling approaches, we propose the \textbf{\ourmethod{}} (\ourmethodshort{}) framework. \ourmethodshort{} is a model-agnostic, simulator-agnostic framework that learns to invoke the costly computational simulator (in a cost-aware manner) as \emph{knowledge-guided correction} to alleviate the effects of rollout errors in surrogates trained with low volumes of training data. Fig.~\ref{fig:hyper-overview} depicts the proposed \ourmethodshort{} framework. Our contributions are as follows.

$\bullet$ \ourmethodshort{} is a first of its kind knowledge-guided correction mechanism that incorporates simulators in the loop without the requirement of the simulators to be \emph{differentiable}.

$\bullet$ \ourmethodshort{} is model agnostic (i.e., functions with any neural surrogate, scientific simulator) and trained in a cost-aware manner, to intelligently invoke the simulator to correct the rollout error of the  neural surrogates.

$\bullet$ We demonstrate through rigorous experimentation on in-distribution, out-of-distribution and noisy data that \ourmethodshort{} significantly reduces rollout error relative to \surrogateonlyshort{} approaches by comparing with state-of-the-art neural surrogates.

\section{Method}
\label{sec:method}

\subsection{PDE Prediction}
\label{sec:problem}

We aim to solve PDEs involving spatial dimensions $\vx =[x_1, x_2, \ldots, x_m] \in \sR^m$ and scalar time $t \in [0, T]$. These PDEs relate solution function $\vu(\vx, t): \sR^m \times [0, T] \rightarrow \sR^n$ to its partial derivatives over the domain. We assume we have initial conditions $\vu(\vx, t=0)$ and boundary conditions $\vu_B(\vx = \vx_B, t)$ which define the field values at time 0 and at the boundaries of the domain respectively. These time-dependent PDEs can be generally defined as:
\small
\begin{equation}
    \frac{\partial \vu} {\partial t} = \mathcal{F}(\vx, t, \vu, \frac{\partial \vu} {\partial \vx}, \frac{\partial^2 \vu} {\partial \vx^2}, \ldots)
\end{equation}
\normalsize
We focus on autoregressive \emph{rollout} of these PDEs over time, which can be defined with an function $g$ taking current state and time as inputs and producing the next state:
\small
\begin{equation}
    g(\vu(\vx, t), \Delta t) = \vu(\vx, t + \Delta t)
\end{equation}
\normalsize
Note that the equation above can be applied autoregressively over any number of timesteps to unroll a PDE trajectory. The cumulative error of this autoregressive process is defined as the \emph{rollout error}, which we aim to minimize.

\subsection{\ourmethodshort{} Components}

\paragraph{Surrogate ML Model} For the sake of conciseness, we shorten $\vu(\vx, t)$ to  $\vu_t$. We begin with a machine learning model $f_{\phi}(\vu_t)$, which we denote as the surrogate. This model can be any deep learning model with parameters $\phi$ that predicts next state $\vu_{t+1}$ given starting state and time $\vu_t$.

\paragraph{Simulator} We also define a PDE simulator $S(\vu_t)$ which numerically solves the PDE to find the next state $\vu_{t + 1}$. Crucially, this simulator is only required to return the next state without any gradient information during training and inference.

\paragraph{Decision Model} Finally, we have a decision model $d_{\theta}(\vu_t)$ which takes current state at time $t$ and outputs a next action: either call the surrogate $f_{\phi}$ or the simulator $S$. In \ourmethodshort{} we implement our decision model as a learned policy $\pi_{\theta}$ which we train using reinforcement learning (RL).

We formalize our decision model using a Markov Decision Process (MDP) which is a tuple $(\sS, \sA, P, r)$. Our states $\sS$ consist of current state $\vu$ and current time $t$. Our action space is binary at each timestep and defined as $a:\{0 = \text{call surrogate}, 1 = \text{call simulator}\}$. Our reward function for a trajectory of length $T$ is:
\small
\begin{equation}
    \label{eq:total-reward}
    R(\va) = \sum_{t=0}^{T-1} - \mathcal{L}(f_{\phi}, S, \vu_t, \vu_{t+1}, a_t) + b(f_\phi, S, \vu_t) - \mathcal{C}(\va, \lambda, T)
\end{equation}
\normalsize

Here $\mathcal{L}$ represents an \emph{error} function, $b$ is a baseline function, and $\mathcal{C}$ is a \emph{cost} function. The baseline function stabilizes the reward values and leads to better policy learning. The baseline function $b$ we use is the mean squared error of randomly calling the simulator the same number of times as our current \ourmethodshort{} policy. Our error function is defined as the mean squared error of either calling the surrogate or simulator according to the policy action:
\small
\begin{equation}
    \label{eq:error-func}
        \mathcal{L}(f_{\phi}, S, \vu_t, \vu_{t+1}, a_t) = \left( \mathcal{G}(\vu_t) - \vu_{t+1} \right)^2
\end{equation}
\begin{equation}
    \mathcal{G}(\vu_t, a_t) = \begin{cases}
      f_\phi(\vu_t) & \text{when } a_t = 0 \\
      S(\vu_t) & \text{when } a_t = 1
   \end{cases}
\end{equation}
\normalsize
$T$ is the length of the trajectory and $\lambda$ is a hyperparameter set by the user to specify what percentage of the trajectory to call the simulator. For example, if $\lambda = 0.5$ then the cost function will penalize the reward if the simulator is not called 50\% of the time. Our cost function is defined as the absolute difference between $\lambda$ and the percent of the trajectory that our policy called the simulator:
\small
\begin{equation}
    \label{eq:cost}
    \mathcal{C}(\va, \lambda, T) = \left| \frac{\| \va \|_1}{T} - \lambda \right|
\end{equation}
\normalsize

Intuitively, reward function \ref{eq:total-reward} optimizes the RL decision model to minimize mean squared error while calling the simulator for $\lambda$ proportion of the trajectory. To learn a policy $\pi_{\theta}(a | \vu_t)$ we use the REINFORCE policy gradient algorithm \citep{sutton1999policy} training with the update:
\small
\begin{equation}
    \nabla_\theta J(\theta) = \mathbb{E}_{\pi_\theta} \left[ \sum_{t=0}^{T-1} \nabla_\theta \log \pi_\theta(a_t \mid \vu_t) \cdot R(\va) \right]
\end{equation}
\normalsize
For the full training algorithm and details of \ourmethodshort{} see Appendix \ref{app:training}.

\subsection{Experiments}\label{sec:experiments}

\textbf{2D Navier Stokes Dataset}. We create a 2D incompressible Navier Stokes fluid flow dataset using $\Phi$Flow \citep{holl2024phiflow} and follow the example of \cite{gupta2022towards}. We use the Navier-Stokes equations in vector velocity form along with an additional scalar field representing particle concentration:
\small
\begin{equation}
    \frac{\partial \vv}{\partial t} = -\vv \cdot \nabla \vv + \mu \nabla^2 \vv - \nabla p + f
    \label{eq:ns_eqn}
\end{equation}
\begin{equation}
    f = \{0, 0.5 c\}
    \label{eq:buoyant_force}
\end{equation}
\begin{equation}
    \nabla \cdot \vv = 0
    \label{eq:mass_cons}
\end{equation}
\begin{equation}
    \frac{\partial c}{\partial t} = - \vv \cdot \nabla c
    \label{eq:concentration_eq}
\end{equation}
\normalsize
Eq.~\ref{eq:ns_eqn} comprises a convection term $-\vv \cdot \nabla \vv$, diffusion term $\mu \nabla^2 \vv$ where $\mu$ indicates kinematic viscosity, a pressure term $\nabla p$, and external force term $f$. Eq. \ref{eq:buoyant_force} defines the buoyant force which is applied in the y-direction. The velocity divergence term Eq. \ref{eq:mass_cons} enforces conservation of mass. Eq. \ref{eq:concentration_eq} models the particle concentration field $c$ that is advected by the velocity vector field $\vv$. Note that $c$ influences $\vv$ through force $f$ and $\vv$ affects $c$ through Eq. \ref{eq:concentration_eq}, creating complex dynamics between fields. For our experiments, we generate 1,000 trajectories of 20 timesteps with $\Delta t = 1.5s$ at a grid size of 64x64, and diffusion coefficient $\mu = 0.01$. We set our velocity boundary condition to $\vv = 0$ (Dirichlet) and our concentration boundary condition to $\partial c / \partial \vx = 0$ (Neumann). While we simulate all the fields above, our experiments focus on predicting the particle concentration field $c$. We split our 1000 trajectories into 3 sets: 400 for surrogate training, 400 for RL training, and 200 for testing. Our RL model selects 4 actions at a time, however \ourmethodshort{} puts no restrictions on action selection.

\textbf{Subsurface Flow Dataset}. To evaluate \ourmethodshort{}'s ability to work with different PDEs and problem scales, we generate a subsurface flow dataset using the Julia-based DPFEHM \citep{pachalieva2022physics} simulator. We use DPFEHM to generate a 2D dataset of subsurface fluid flows modeled by the Richards equation:.
\small
\begin{equation}
    \frac{\partial \theta}{\partial t} = \nabla \cdot \mathbf{K}(h) (\nabla h + \nabla z) - T^{-1}
    \label{eq:richards}
\end{equation}
\normalsize
Eq. \ref{eq:richards} models fluid flow underground in an unsaturated medium. $\theta$ represents the volumetric fluid content, $\mathbf{K}(h)$ is the unsaturated hydraulic conductivity, $h$ is the pressure, $\nabla z$ is the geodetic head gradient, and $T^{-1}$ is the fluid sink term. We generate 500 trajectories of 100 timesteps each with a grid size of 50x50 and timestep size of 10 seconds. Out of the 500 trajectories, 200 are used for surrogate training, 200 are used for RL training, and 100 are used for testing. We use Dirchlet boundary conditions (0) for volumetric flux on all sides of our domain but inject fluid at the top center of the domain at a rate of $0.01 m/s$. In our experiments, we predict the pressure field $h$.

\section{Results and Discussion}
In this section, we investigate the effectiveness of \ourmethodshort{} to model transient PDE systems efficiently and with minimal rollout error compared to surrogate-only (\surrogateonlyshort{}) rollout. We compare with state-of-the-art neural surrogates including \unet{} and Fourier Neural Operator (\fno{}), \mpp{}, and \refiner{}. Specifically, we investigate the effectiveness of \ourmethodshort{} rollouts under changing PDE dynamics and in noisy data settings. Further, we also demonstrate the surrogate-agnostic and simulator-agnostic nature of \ourmethodshort{}. 
Our experiments seek to answer the following research questions:

$\bullet$ \textbf{RQ1:} How effective are \ourmethodshort{} rollouts compared to \surrogateonlyshort{} rollouts for transient PDE systems?

$\bullet$ \textbf{RQ2:} Are \ourmethodshort{} rollouts effective under changed physical conditions? 

$\bullet$ \textbf{RQ3:} Are \ourmethodshort{} rollouts effective under noisy data conditions?

$\bullet$ \textbf{RQ4:} How crucial is the intelligent decision model for effective \ourmethodshort{} rollouts?

$\bullet$ \textbf{RQ5:} What is the error/efficiency trade-off between \surrogateonlyshort, \ourmethodshort{}, and simulator-only paradigms?

\subsection{RQ1: \ourmethodshort{} rollouts vs \surrogateonlyshort}

We begin by comparing \ourmethodshort{} to six \surrogateonlyshort{} baselines: \emph{\unet{}}, \emph{\fno{}}, \emph{\unetmultistep}, \emph{\mppzero}, \emph{\mpp}, and \emph{\refiner}. All baselines except \unetmultistep{} are trained in a one-step-ahead manner i.e., given the current state of the resolved field they are trained to predict the next state. Mean squared error (MSE) loss is used to train all surrogates over the full resolution trajectory. Note that all these baselines except \mppzero{} are trained with the same dataset that HyPER is trained with. For full training details see Appendix \ref{app:training} and for baseline details see Appendix \ref{app:surr-models}.

\begin{table}[!h]
\caption{2D Navier-Stokes results with best results in bold. Notice that \ourmethodshort{} rollout incurs significantly lower rollout (cumulative) error compared to \surrogateonlyshort{} models.}
\label{tab:ns}
\begin{center}
  \small
  \begin{adjustbox}{max width=\textwidth}
  \begin{tabular}{l|ccccccc}
    \toprule
    \textbf{2D Navier-Stokes}
      & \unet & \fno & \unetmultistep & \mppzero & \mpp & \refiner & \ourmethodshort \\
      \midrule
    Final MSE & 0.019 & 0.053 & 0.016 & 0.22 & 0.036 & 0.024 & \textbf{0.011} \\
    Cumulative MSE & 0.312 & 0.882 & 0.311 & 4.495 & 0.747 & 0.405 & \textbf{0.164} \\
    \bottomrule
  \end{tabular}
  \end{adjustbox}
\end{center}
\end{table}

\begin{figure}[!t]
\begin{center}
\includegraphics[width=0.85\linewidth]{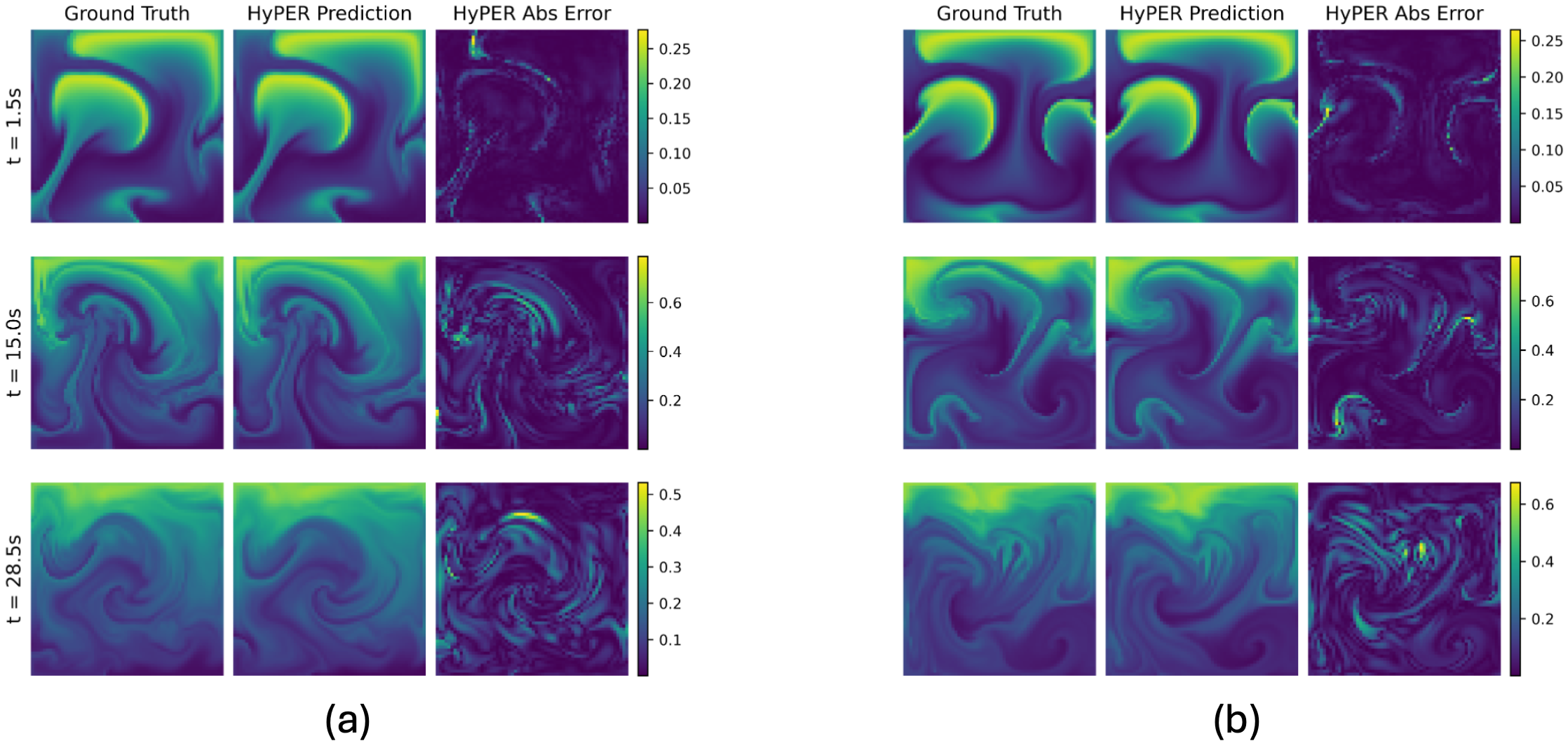}
\end{center}
\caption{Predictions and absolute error snapshots of \ourmethodshort{} rollout for two distinct trajectories.}
\vspace{-2ex}
\label{fig:clean-ns-preds}
\end{figure}

The \unet{} model is based on the modern UNet architecture in PDEArena \citep{gupta2022towards} while the \fno{} model is built with the \emph{neuraloperator} library \citep{kovachki2021neural,li2020fno}. \unetmultistep{} is trained to specialize for rollout trajectory prediction, i.e. it minimizes MSE for 20-step autoregressive prediction. \mppzero{} (zero-shot) and \mpp{} \citep{mccabe2023multiple} are large transformer based multi-task models which have been pretrained on a variety of fluid prediction tasks. \mppzero{} is the off-the-shelf pretrained model, while \mpp{} is finetuned on our data. \refiner{} \citep{lippe2024pde} is a modern diffusion based model which both predicts the next timestep while adding a multi-step noising/denoising process and has been shown to reduce rollout error. \ourmethodshort{} utilizes a \unet{} model that is pretrained on 400 trajectories while the RL model of \ourmethodshort{} is trained on a separate set of 400 trajectories with $\lambda = 0.3$. All \surrogateonlyshort{} baselines are trained using 800 trajectories (same as \ourmethodshort{} RL plus surrogate dataset) for 200 epochs. Table \ref{tab:ns} shows the MSE of the final trajectory state (Final MSE) as well as the aggregated MSE (Cumulative MSE) over all trajectory states in a rollout. Both metrics are calculated as an average across 200 test trajectories for both \surrogateonlyshort{}  and \ourmethodshort{} rollouts. \ourmethodshort{} outperforms all baselines significantly, with an average improvement in cumulative rollout error of \textbf{68.30\%}. While \unetmultistep{} seems to be the best performing \surrogateonlyshort{} method, because it trains across the full trajectory it learns a smoothed prediction that has poor high frequency detail (suffering from spectral bias) in comparison to \unet{} (see \ref{app:unet-multistep-predictions} for examples). By incorporating the simulator, \ourmethodshort{} yields significantly lower cumulative rollout error while only invoking the simulator during $\sim 30\%$ of the trajectory for knowledge-guided correction of surrogate rollout.

In Figure \ref{fig:ns-cum-mse-surr-rl} we plot the rollout MSE performance of \ourmethodshort{} versus the surrogate only (\surrogateonlyshort{}) methods for a single sample trajectory. Note that \ourmethodshort{} outperforms both \surrogateonlyshort{} models by a large margin while only invoking the simulator six times in the 20-step trajectory. Also see Figures \ref{fig:clean-ns-preds}(a) and \ref{fig:clean-ns-preds}(b) which depict qualitative snapshots from two distinct \ourmethodshort{} rollouts.

\subsection{RQ2: \ourmethodshort{} rollouts with Changing Physical Conditions}
\vspace{-1ex}
Pretrained neural surrogates are often confronted with modeling contexts with similar PDE dynamics but varied initial / boundary conditions. Effective rollouts under changing physical conditions is hence a crucial requirement for the effective use of neural surrogates in computational science. To this end, we inspect rollout errors of \surrogateonlyshort{} and \ourmethodshort{}, under changing boundary conditions. Specifically, to test the adaptability of \ourmethodshort{}, we investigate \ourmethodshort{} rollouts with changing boundary conditions in our Navier-Stokes experiment.  We generate a separate Navier-Stokes dataset which follows the same initial conditions of our previous experiment, but comprises a different velocity boundary condition at the top boundary of the domain. Specifically, the velocity at the top boundary of the domain is increased from 0.0 to 0.5 m/s (imposing an intermittent external forcing effect causing the fluid to escape from the top of the domain) for four intermediate time-steps of the trajectory (timesteps 12-16). Following this, the boundary condition is reverted back to 0.0 for the rest of the trajectory. This results in the fluid escaping out of the top during these time steps changing the PDE dynamics significantly.

\begin{table}[!h]
\caption{Results depict rollout error for 2D Navier-Stokes with changing boundary conditions. Notice that \ourmethodshort{} accumulates lower rollout error compared to \surrogateonlyshort{} approaches.}
\label{tab:ns-boundary}
\begin{center}
\small
  \begin{adjustbox}{max width=\textwidth}
  \begin{tabular}{l|ccccc}
    \toprule
    \textbf{Changing Boundary}
      & {\unetonly} & {\fnoonly} & \unet & \fno & {\ourmethodshort} \\
      \midrule
    Final MSE & 0.362 & 0.412 & \textbf{0.014} & 0.027 & 0.027 \\ 
    Cumulative MSE & 1.84 & 2.538 & 0.92 & 0.969 & \textbf{0.527} \\
    \bottomrule
  \end{tabular}
  \end{adjustbox}
\end{center}
\vspace{-2ex}
\end{table}

\begin{figure}[h]
\begin{center}
\includegraphics[width=0.91\linewidth]{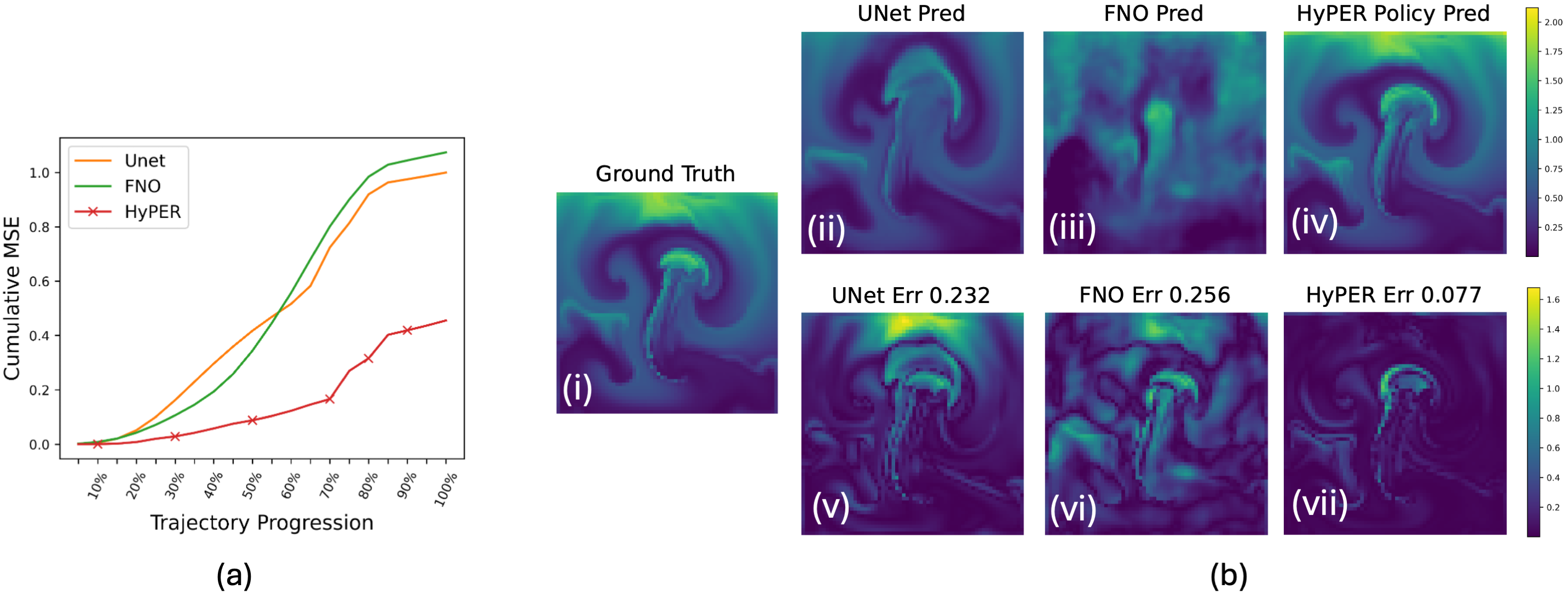}
\vspace{-1ex}
\end{center}
\caption{Predictions and absolute error of \ourmethodshort{} vs \unet{} and \fno{} for a single trajectory and timestep. Fig.~\ref{fig:boundary-qual-error}a shows the rollout error accumulation over the trajectory with `x' marking the times the simulator is called. Fig.~\ref{fig:boundary-qual-error}b shows the resolved system state for a the same trajectory at a single timestep. Fig.~\ref{fig:boundary-qual-error}b(i) shows the ground truth field while Fig.~\ref{fig:boundary-qual-error}b(ii)-(iv) indicate the result of \unet{}, \fno{} and \ourmethodshort{} rollouts respectively. Fig.~\ref{fig:boundary-qual-error}b(v)-(vii) depict the corresponding absolute errors. We notice that only \ourmethodshort{} rollouts capture the correct characteristics relative to the ground truth owing to the knowledge-guided correction while \surrogateonlyshort{} models are unable to faithfully resolve the trajectory under changing physical conditions.}
\label{fig:boundary-qual-error}
\vspace{-3ex}
\end{figure}

In this case, surrogates \unetonly{} and \fnoonly{} are not re-trained with the changed boundary condition and hence show poor performance accumulating significant rollout error. \unet{} and \fno{} are trained on the changing boundary data (the same dataset as \ourmethodshort{}), but still suffer from poor rollout error.
We train our \ourmethodshort{} RL policy with a pretrained surrogate (not trained on to the changed boundary setting) and a simulator (fully aware of the changed boundary condition). The RL policy of \ourmethodshort{} is trained with 400 of these changed trajectories to learn the optimal policy of (frugally) invoking the simulator to apply knowledge-guided correction to reduce surrogate rollout errors.
As shown in Table \ref{tab:ns-boundary}, \ourmethodshort{} reduces cumulative error in this scenario by \textbf{71.36\%} and \textbf{79.20\%} relative to~\unetonly{} and~\fnoonly{} rollout errors respectively. Even when compared to surrogates trained on the changing boundary data \unet{} and \fno{}, \ourmethodshort{} reduces rollout error by  \textbf{42.72\%} and \textbf{45.61\%} respectively. In Figure~\ref{fig:boundary-qual-error}a we demonstrate that \ourmethodshort{} has much lower rollout error than the~\unetonly{} and~\fnoonly{} models. Figure~\ref{fig:boundary-qual-error}b shows sample qualitative predictions of \ourmethodshort{} and \surrogateonlyshort{} rollouts under the changing boundary scenario of interest, to further reinforce our point. As illustrated, our model prediction is much closer to the ground truth after the boundary condition has changed and fluid has escaped the box owing to the appropriately invoked knowledge-guided correction by the learned RL policy in response to increasing surrogate rollout error. Fig.~\ref{fig:boundary-qual-error}a shows the significantly lower rollout error for \ourmethodshort{} rollout relative to \unetonly{} and \fnoonly{} models.

\subsection{RQ3: \ourmethodshort{} rollouts with Noisy Data}

Neural surrogates, in addition to being data hungry and accumulating rollout error, have also been known to exhibit catastrophic failures in challenging PDE conditions (e.g., stiff PDEs)~\cite{krishnapriyan2021characterizing}. As neural surrogates are a crucial part of \ourmethodshort{} rollouts, it is imperative to investigate whether  \ourmethodshort{} is capable of \emph{adapting} to such failures by invoking the knowledge-guided correction (i.e., the simulator) to minimize the propagation (and accumulation) of such local failures over the remaining trajectory rollout. Further, another crucial property to investigate is the robustness of the intelligent decision mechanism in \ourmethodshort{}, to noisy, low-quality surrogate predictions.
To jointly investigate both goals, we consider a simple (contrived) context with noisy inputs supplied to the \ourmethodshort{} RL policy. This additive noise, injected at specific steps to \emph{corrupt} the surrogate output in the trajectory, serves to mimic low-quality surrogate predictions. Hence, experiments with such noisy inputs help characterize the ability of \ourmethodshort{} to adapt to sudden local changes during rollout (like catastrophic surrogate failure) in addition to demonstrating its ability for robust decision-making under noisy data conditions.

To carry out this investigation, we add random Gaussian noise with mean 0 at four separate variance ($\sigma^2$) scales at fixed timesteps of our trajectory. The \unetonly{} and \fnoonly{} surrogate models are never trained with this noisy data and therefore perform poorly when encountering it. The \unet{} and \fno{} models are trained with the noisy data and show improvement. We train the RL policy of \ourmethodshort{} with a small set (400) of these noisy trajectories while keeping our surrogate model static (using the \unetonly{} model). We test two different noise corruption scenarios added to a 20 step trajectory rollout. (i) unimodal noise: a case where noise is added at timesteps [12-16] and (ii) bimodal noise: a more sophisticated case where noise is added at two different time windows of [2-4] and [15-16]. The unimodal noise results are presented in Table \ref{tab:ns-noise-single-mode}, where we see that \ourmethodshort{} rollouts outperform state-of-the-art \surrogateonlyshort{} approaches. In this case we notice a reduction in cumulative MSE by \textbf{19.81}\%-\textbf{49.90}\% across the four noise scales. Note that \ourmethodshort{} achieves this significant reduction in rollout error while using the un-specialized \unetonly{} surrogate, demonstrating that \ourmethodshort{} succeeds in adapting pretrained models to noisy data.

\begin{table}[t]
\caption{Results depicting rollout error for the 2D-Navier-Stokes experiment, with random Gaussian noise added to inputs, at timesteps 12-16. $\sigma^2$ is the \emph{scale} of the noise. The percentage reduction of cumulative MSE rollout error by \ourmethodshort{}, over the best performing \surrogateonlyshort{} model is in parentheses.}
\label{tab:ns-noise-single-mode}
\begin{center}
\small 
\setlength{\tabcolsep}{3pt}
  \begin{tabular}{l|cccccc}
    \toprule
    \multirow{2}{*}{\textbf{Experiment}} &
      \multicolumn{5}{c}{\textbf{Cumulative MSE}} \\
      & {\unetonly} & {\fnoonly} & {\unet} & {\fno} & {\ourmethodshort} \\
      \midrule
    Unimodal $\sigma^2=1.0$ & 2.931 & 2.523 & 1.61 & 1.713 & \textbf{1.291} (19.81\%) \\
    Unimodal $\sigma^2=0.75$ & 1.746 & 1.916 & 1.013 & 1.291 & \textbf{0.8} (21.03\%) \\
    Unimodal $\sigma^2=0.5$ & 1.009 & 1.487 & 0.642 & 1.137 & \textbf{0.49} (23.68\%) \\
    Unimodal $\sigma^2=0.25$ & 0.612 & 1.246 & 0.519 & 0.862 & \textbf{0.26} (49.90\%) \\
    \bottomrule
  \end{tabular}
\end{center}
\vspace{-1ex}
\end{table}

\subsection{RQ4: Investigating \ourmethodshort{} Rollouts vs. Random Policy Rollouts}
To evaluate whether \ourmethodshort{} learns an effective RL policy, we compare it to a \randompol{} baseline. This baseline is designed to invoke the simulator the same number of times as the \ourmethodshort{} RL policy for a particular rollout, but with uniform random probability over each timestep. By comparing the \ourmethodshort{} policy rollout to a \randompol{} rollout with the same budget, we show that \ourmethodshort{} learns a superior performing policy in our experimental scenarios.
\begin{table}[!ht]
\caption{\ourmethodshort{} versus a random policy which calls the simulator the same number of times.}
\label{tab:ns-random-policy}
\begin{center}
\small 
\setlength{\tabcolsep}{3pt}
  \begin{tabular}{l|cc|cc|c}
    \toprule
    \multirow{2}{*}{\textbf{Experiment}} &
      \multicolumn{2}{c|}{\textbf{Final MSE}} &
      \multicolumn{2}{c}{\textbf{Cumulative MSE}} &
      \multicolumn{1}{c}{\textbf{Cumulative Wins \%}} \\
      & {\randompol} & {\ourmethodshort} & {\randompol} & {\ourmethodshort} & {\ourmethodshort} \\
      \midrule
    Noise Free & \textbf{0.011} & \textbf{0.011} & 0.186 & \textbf{0.164} & 57.00\% \\
    Bimodal $\sigma^2=1.0$ & 0.121 & \textbf{0.07} & 2.443 & \textbf{1.706} & 73.00\% \\
    Bimodal $\sigma^2=0.75$ & 0.055 & \textbf{0.025} & 1.305 & \textbf{0.739} & 84.50\% \\
    Bimodal $\sigma^2=0.5$ & 0.037 & \textbf{0.027} & 0.891 & \textbf{0.683} & 67.00\% \\
    Bimodal $\sigma^2=0.25$ & 0.022 & \textbf{0.019} & 0.495 & \textbf{0.424} & 61.00\% \\
    Changing Boundary & 0.149 & \textbf{0.027} & 0.865 & \textbf{0.527} & 80.50\% \\
    \bottomrule
  \end{tabular}
\end{center}
\vspace{-2ex}
\end{table}

We summarize these results in Table \ref{tab:ns-random-policy}, which also shows the percentage of trajectories over which \ourmethodshort{} reduces MSE compared to \randompol{} (Cumulative Wins \%). A `win' is characterized by a \ourmethodshort{} rollout that yields a lower cumulative MSE than the corresponding \randompol{} rollout. In the `Noise Free' trajectories, \ourmethodshort{} only wins for 57\% of the trajectories because the MSEs at each timestep in the Noise Free case have low variance, so a uniform \randompol{} performs fairly well. However, in the bimodal noise scales (with fast and large rollout error accumulation), precise invocation of the simulator for knowledge-guided correction is imperative to prevent error accumulation. Hence, we see \ourmethodshort{} owing to its intelligent (RL-based) decision policy, has a higher percentage of wins ($\approx 70\%$) and significantly lower cumulative MSE at higher noise scales. We also see the strength of \ourmethodshort{}'s learned policy when considering the changing boundary condition experiment which has a 80.50\% win rate over the \randompol{} and reduces cumulative MSE by 39.08\%. This result demonstrates the effectiveness of \ourmethodshort{} in realistic physical scenarios and noisy conditions.

\subsection{RQ5: Cost versus Accuracy Trade-Off}

\begin{figure}[h]
\begin{center}
\includegraphics[width=0.70\linewidth]{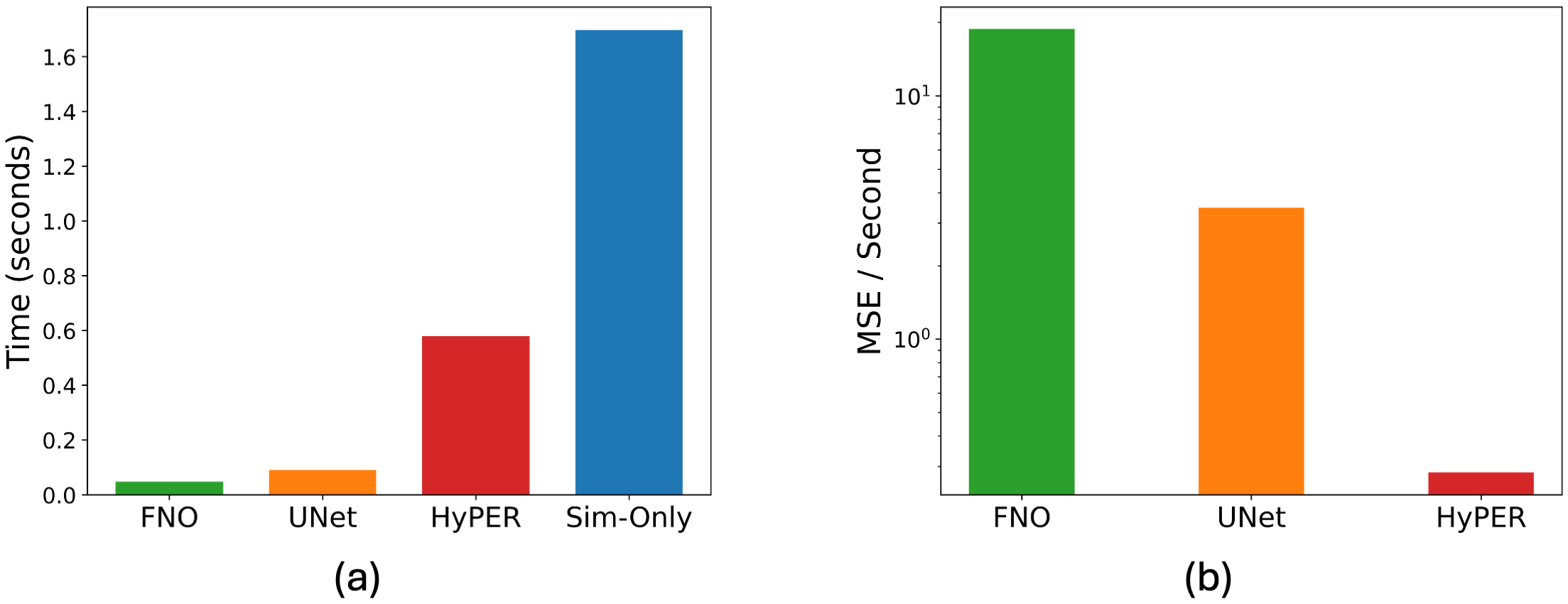}
\end{center}
\vspace{-1.5ex}
\caption{Figure \ref{fig:time-bar-charts}(a) shows the average time of PDE prediction of a full trajectory for each method. Figure \ref{fig:time-bar-charts}(b) illustrates the error per unit time (\emph{lower is better}) for each method. We do not show the Sim-Only case here as we assume error is effectively zero for the simulator.}
\label{fig:time-bar-charts}
\vspace{-1.5ex}
\end{figure}

A natural question is how much of a cost we pay in wall clock time when utilizing \ourmethodshort{} compared to baselines? We evaluate this by measuring the average time of each method on our 200 test trajectories and show results in Figure \ref{fig:time-bar-charts}(a). While the wall clock time of \surrogateonlyshort{} methods is lower than \ourmethodshort{}, our method has much lower rollout error, which we illustrate in Figure \ref{fig:time-bar-charts}(b). Here we show the error per unit time (\emph{lower is better}) and we see that \ourmethodshort{} has lower rollout error per second by a large margin in comparison to the baselines (note this plot is on a log scale). While calling the simulator incurs a time cost, the user can specify the $\lambda$ parameter to adjust to their requirements allowing \ourmethodshort{} to be flexible in varying settings.

\subsection{Surrogate-Agnostic and Simulator-Agnostic Design of \ourmethodshort{}}

To demonstrate the model-agnostic capability of \ourmethodshort{}, we train it on the subsurface flow task (dataset details in Sec.~\ref{sec:experiments}), comprising longer (i.e., 100 step) trajectories. In this case, we train \ourmethodshort{} using the Julia simulator DPFEHM, which simulates underground fluid flow in porous media. 
\begin{table}[!h]
\caption{2D Subsurface flow experiment results for a 100 timestep trajectory.}
\label{tab:subsuface}
\begin{center}
\small
  \begin{tabular}{l|ccc}
    \toprule
    \multirow{2}{*}{\textbf{Experiment}} &
      \multicolumn{3}{c}{\textbf{Cumulative MSE per timestep}} \\
    & {\unetonly} & {\fnoonly} & {\ourmethodshort} \\
      \midrule
    Subsurface & 4.345 & 10.938 & \textbf{0.271} \\
    \bottomrule
  \end{tabular}
\end{center}
\end{table}
As we see in Table \ref{tab:subsuface}, \ourmethodshort{} outperforms both \surrogateonlyshort{} baselines by a cumulative MSE per timestep reduction of 93.76\% and 97.52\%. The integration of a distinct Julia-based simulator in a scenario with larger physical scales and times, shows that \ourmethodshort{} is surrogate and simulator agnostic and can perform well and reduce rollout error in multiple problem settings.

\section{Related Work}
\label{sec:related_work}
\vspace{-1ex}

\textbf{\surrogateonlyshort{}} approaches circumvent the use of computational simulations during inference and only employ simulations to generate training data.
The U-Net~\citep{ronneberger2015u} model is a popular approach owing to its ability to capture spatial and temporal dynamics at multiple scales. This model has recently~\citep{gupta2022towards} demonstrated state of the art performance on various fluid dynamics benchmarks. Operator learning~\citep{kovachki2021neural} approaches that learn function families of PDEs rather than single PDE instances have also been investigated to be effective neural surrogates. Two notable operator learning models are the deep operator network~\citep{lu2021learningdeeponet} and the Fourier neural operator (FNO)~\citep{li2020fno} models. Multiple investigations have been carried out employing operator learning techniques including extending them to multi-resolution~\citep{howard2023multifidelitydatahungry,lu2022multifidelitydatahungry} settings. Recently ~\cite{takamoto2022pdebench} have also demonstrated that FNOs yield state-of-the-art results on benchmark tasks. 

\textbf{Knowledge-guided \surrogateonlyshort{}} approaches like the popular Physics-informed neural network (PINN)~\citep{raissi2019physics,jagtap2020extended,cuomo2022scientific} have also been effectively employed for improved generalization. A related paradigm of Universal Differential Equations~\citep{rackauckas2020universal} utilizes data-driven surrogates to estimate focused sub-components of governing equations for better process modeling. Such approaches assume end-to-end gradient based training in a physics-informed manner and are known to converge slowly and to trivial solutions under data paucity and stiff PDE conditions~\cite{krishnapriyan2021characterizing} owing to catastrophic gradient imbalance~\cite{wang2021understanding} between data-driven and physics-guided loss terms.

\textbf{\surrogateonlyshort{} for Transient PDE Dynamics.} All \surrogateonlyshort{} approaches struggle to model transient PDE systems in an autoregressive manner and incur rollout error~\citep{carey2024data}. In the work of ~\cite{lippe2024pde}, it is demonstrated how the spectral bias of traditional neural surrogates leads to significant error accumulation and they propose an initial \emph{refinement} solution inspired by the process in diffusion modeling.
Separately,~\cite{list2024temporal} have conducted a characterization of \surrogateonlyshort{} rollout error and comment about the significant improvement obtainable by incorporating simulators in-the-loop.

\textbf{Hybrid-modeling for Transient PDE Dynamics.} Hybrid-modeling approaches retain the simulation and resolve each query employing the neural surrogate and the simulator `in-the-loop'. One recent example is \cite{zhang2022hybrid} which selects between a simulator and a surrogate (in a pre-defined rule-based maner) to resolve a PDE trajectory. In their work, the simulator and surrogate are each invoked a pre-fixed number of times. In contract, \ourmethodshort{} learns a dynamic policy, capable of adapting to scenarios such as changing boundary conditions. Another major drawback with many such approaches~\citep{chen2018neural,belbute2020combining,um2020solver,donti2021dc3,pachalieva2022physics} is the requirement of simulators to be \emph{differentiable} as they are mostly employed as additional layers in the neural network architecture, to be trained end-to-end with the neural surrogates.

\vspace{-1ex}
\section{Conclusion and Future Work}
\vspace{-1.5ex}
This work presents a first of its kind knowledge-guided correction mechanism to reduce rollout errors in neural surrogates that model transient PDE systems. Our proposed method \ourmethod{} (\ourmethodshort{}) learns a reinforcement learning based cost-aware control policy to parsimoniously invoke (costly) simulation steps to \emph{correct} erroneous surrogate predictions. In contrast to existing approaches that employ simulators `in-the-loop` with neural surrogates, \ourmethodshort{} does not impose any differentiability restrictions on the computational simulations. Further, \ourmethodshort{} is surrogate and simulator agnostic and is applicable to any neural surrogate and off-the-shelf simulator capable of resolving transient PDE systems.

We have demonstrated the effectiveness of our proposed \ourmethodshort{} model in traditional in-distribution rollouts, under changing physical conditions and under noisy data conditions. Overall \ourmethodshort{} yields significant improvements of cumulative rollout error over state-of-the-art surrogate-only approaches with an average \textbf{68.30}\% improvement for in-distribution rollouts, \textbf{75.28}\% improvement for rollouts under changing physical conditions and \textbf{28.61}\% improvement for rollouts under noisy data conditions.
In the future, we will investigate more sophisticated actor-critic based RL policies based to further improve the sample efficiency of \ourmethodshort{}. We will also explore extensions of \ourmethodshort{} to more challenging problems in multi-physics contexts as well as multi-phase flows. 

\subsubsection*{Acknowledgments}
DO gratefully acknowledges support from the Department of Energy, Office of Science, Office of Basic Energy Sciences, Geoscience Research program under Award Number LANLECA1.

\bibliography{refs}
\bibliographystyle{iclr2025_conference}

\clearpage
\appendix
\section{Appendix}

\subsection{\ourmethodshort{} Training Procedure}
\label{app:training}

Below we detail the RL training procedure of \ourmethodshort{}.

\begin{algorithm}
  \caption{HyPER Training Algorithm
    \label{alg:hyper-training}}
  \begin{algorithmic}[1]
    \Require{Dataset: $\mathcal{D}$, Pretrained Surrogate Model: $f_{\phi}$, Simulator: $S$, Decision Model: $d_{\theta}$, \newline Trajectory length: $\tau$, Simulator proportion hyperparameter: $\lambda$, Learning rate: $\eta$, \newline Error function (MSE in our case): $\ell$, Cost function: $c$}
    \For{$\vu, \vz \in \mathcal{D}$} \Comment{For every trajectory in dataset, get field $\vu$ and conditional features $\vz$}
      \State $k \gets 0$ \Comment{Initialize number of simulator calls}
      \State $R_d \gets [\ ]$ \Comment{Initialize decision model rewards list} \vspace{2pt}
      \State $R_b \gets [\ ]$ \Comment{Initialize baseline rewards list} \vspace{2pt}
      \State $L \gets [\ ]$ \Comment{List to store log probabilities of each action}
      \State $\hat{\vu}(\vx, -1) \gets 0$ \Comment{Set initial field prediction to 0}
      \For{$t \in [0, \tau]$} \Comment{For every time-step in trajectory}
        \State $p \gets d_{\theta}(\hat{\vu}(\vx, t-1), \vz, t)$ \Comment{Get RL model probabilities for next action}
        \State $a \sim \mathrm{Bernoulli}(p)$ \Comment{Sample next action}
        \State $L \pluseq a \log(p) + (1-a) \log(1-p)$ \Comment{Store log probability of action}
        \If{$a = 0$}
          \State $\hat{\vu}(\vx, t) \gets f_{\phi}(\hat{\vu}(\vx, t-1), t)$ \Comment{Call surrogate for next step prediction}
        \ElsIf{$a = 1$}
          \State $\hat{\vu}(\vx, t) \gets S(\hat{\vu}(\vx, t-1), t)$ \Comment{Call simulator for next step prediction}
          \State $k \pluseq 1$ \Comment{Track number of times simulator called}
        \EndIf
        \State $R_d \pluseq -\ell(\hat{\vu}(\vx, t), \vu(\vx, t)) - c(\lambda, k, \tau)$ \parbox[t]{.45\linewidth}{\Comment{Store policy reward based on MSE and cost function}}
      \EndFor
      \State $I \sim \mathrm{Uniform Without Replacement}([0, \tau], k)$ \parbox[t]{.45\linewidth}{\Comment{Sample $k$ times between $[0, \tau]$ without replacement, this list will contain the time-steps at which the random baseline will call the simulator}} \vspace{4pt}
      \For{$t \in [0, \tau]$} \Comment{Run random baseline for trajectory}
        \If{$t \notin I$}
          \State $\hat{\vu}(\vx, t) \gets f_{\phi}(\hat{\vu}(\vx, t-1), t)$ \Comment{Call surrogate for next step prediction}
        \ElsIf{$t \in I$}
        \State $\hat{\vu}(\vx, t) \gets S(\hat{\vu}(\vx, t-1), t)$ \Comment{Call simulator for next step prediction}
        \EndIf
        \State $R_b \pluseq -\ell(\hat{\vu}(\vx, t), \vu(\vx, t)) - c(\lambda, k, \tau)$ \parbox[t]{.4\linewidth}{\Comment{Store baseline reward using MSE and cost function}}
      \EndFor
      \State $\nabla_{\theta} J \gets - \nabla_{\theta} L \cdot \mathrm{StopGradient}(R_d - R_b)$ \parbox[t]{.44\linewidth}{\Comment{Calculate policy gradient, this is done element-wise and then summed}} \vspace{4pt}
      \State $\theta \gets \theta - \eta \nabla_{\theta} J$ \Comment{Update RL decision model parameters}
    \EndFor
  \end{algorithmic}
\end{algorithm}

\subsection{RQ3: \ourmethodshort{} Rollouts with Noisy Data: Bimodal Noise Experiments}\label{sec:multimodalnoise}

\begin{table}[!h]
\caption{Results depicting rollout error for the 2D-Navier-Stokes experiment with random Gaussian noise added at timesteps 2-4, 15-16. $\sigma^2$ is the \emph{scale} of the noise. The percentage reduction of cumulative MSE rollout error by \ourmethodshort{}, over the best performing \surrogateonlyshort{} model is in parentheses.}
\label{tab:ns-noise-multi-mode}
\begin{center}
\small
\setlength{\tabcolsep}{3pt}
  \begin{tabular}{l|ccc|ccc}
    \toprule
    \multirow{2}{*}{\textbf{Experiment}} &
      \multicolumn{3}{c|}{\textbf{Final MSE}} &
      \multicolumn{3}{c}{\textbf{Cumulative MSE}} \\
      & {\unetonly} & {\fnoonly} & {\ourmethodshort} & {\unetonly} & {\fnoonly} & {\ourmethodshort} \\
      \midrule
    Bimodal $\sigma^2=1.0$ & 0.844 & 0.097 & \textbf{0.07} & 6.648 & 2.914 & \textbf{1.706} (41.46\%) \\
    Bimodal $\sigma^2=0.75$ & 0.25 & 0.079 & \textbf{0.025} & 3.436 & 2.126 & \textbf{0.739} (65.24\%) \\
    Bimodal $\sigma^2=0.5$ & 0.087 & 0.07 & \textbf{0.027} & 1.942 & 1.576 & \textbf{0.683} (56.67\%) \\
    Bimodal $\sigma^2=0.25$ & 0.046 & 0.068 & \textbf{0.019} & 1.095 & 1.266 & \textbf{0.424} (61.28\%) \\
    \bottomrule
  \end{tabular}
\end{center}
\end{table}

Table \ref{tab:ns-noise-multi-mode} depicts the performance of \ourmethodshort{} in the bimodal noise case.
We notice results similar  to the unimodal noise case, with a reduction in cumulative MSE error of \textbf{41.46}\%-\textbf{65.24}\%. We believe this improved performance in a more complex noise distribution is a result of our RL policy learning the more complex error distribution while the \surrogateonlyshort{} methods accumulate larger error over two different noise windows.

\subsection{\ourmethodshort{} Average Performance}

\begin{figure}[h]
\begin{center}
\includegraphics[width=0.5\linewidth]{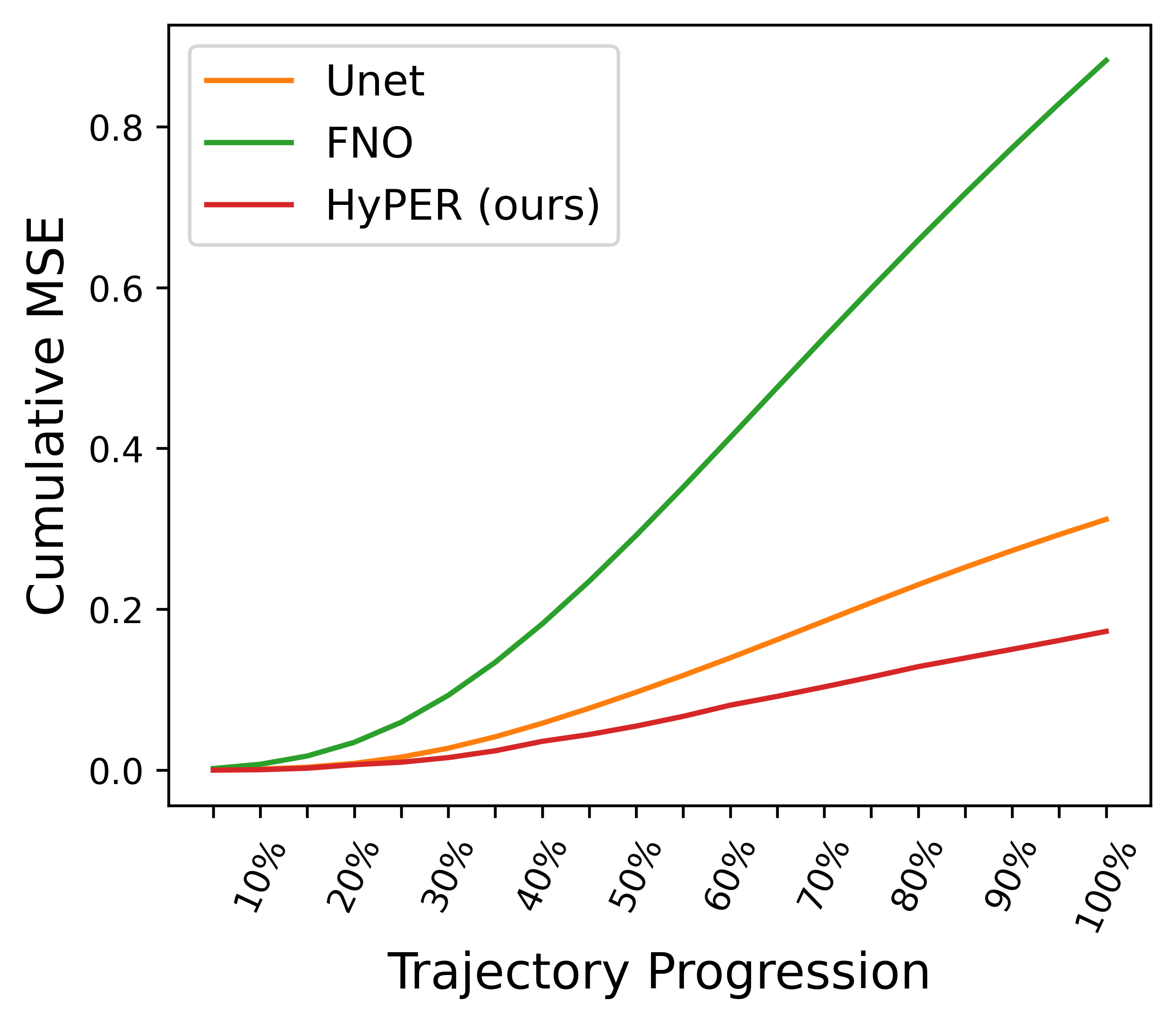}
\end{center}
\caption{Average Cumulative MSE over all test trajectories of \ourmethodshort{} vs. \unet{} and \fno{}.}
\label{fig:ns-cum-mse-avg}
\end{figure}

Figure \ref{fig:ns-cum-mse-avg} shows the average rollout error of \ourmethodshort{} versus \unet{} and \fno{} methods for all 200 Navier-Stokes test trajectories. This demonstrates that \ourmethodshort{} is effective in reducing rollout error significantly over a large set of unseen test trajectories.

\subsection{\unetmultistep{} Prediction Quality}
\label{app:unet-multistep-predictions}

\begin{figure}[h]
\begin{center}
\includegraphics{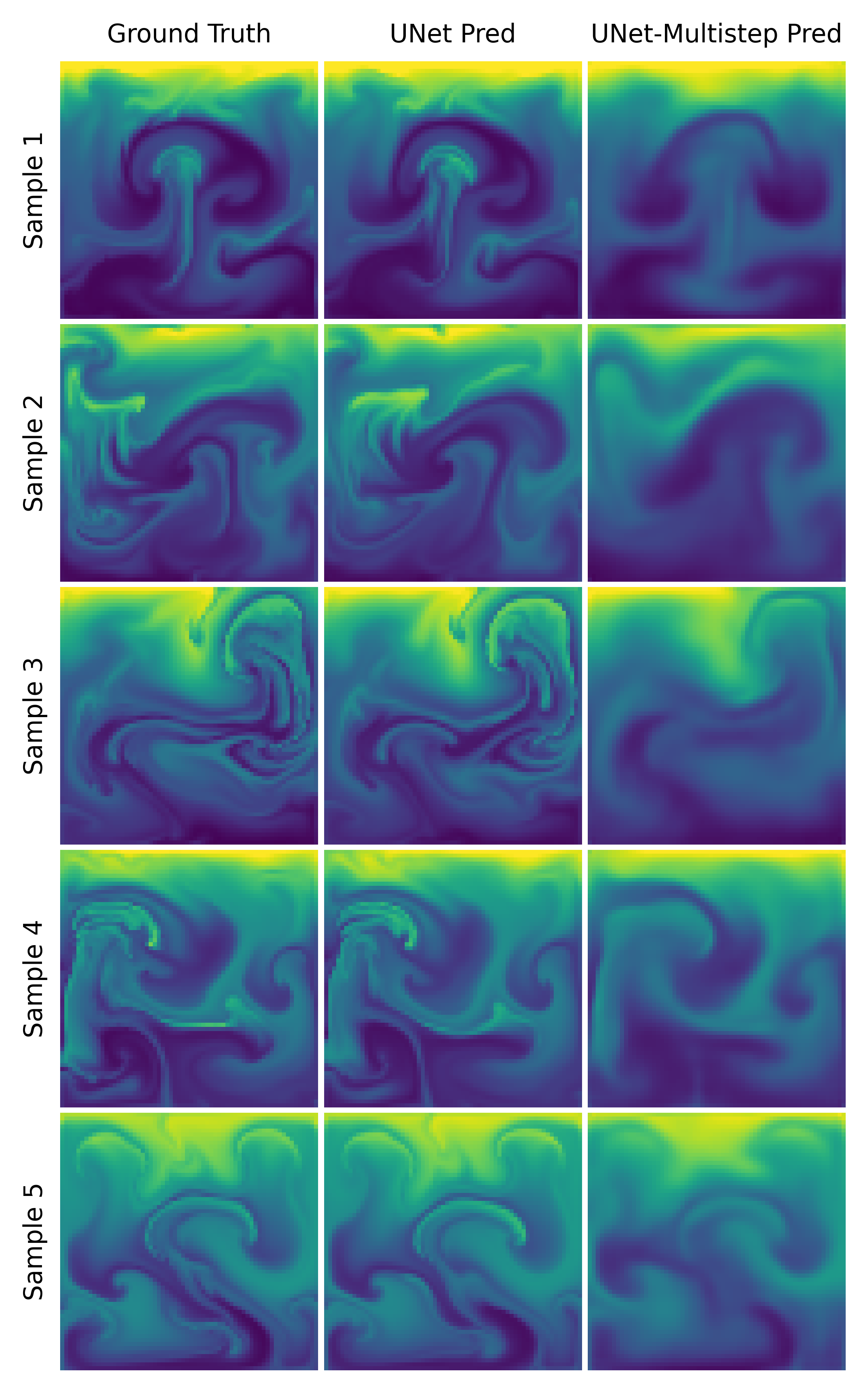}
\end{center}
\caption{Sample predictions of \unet{} vs \unetmultistep{} on five different noise free trajectories. Note that \unetmultistep{} predictions suffer from spectral bias and have poor high frequency detail.}
\label{fig:multistep-unet-vs-unet}
\end{figure}

In Figure \ref{fig:multistep-unet-vs-unet} below we show five different sample predictions of \unet{} versus \unetmultistep{}. Our investigations revealed that in comparison to \unet{}, all \unetmultistep{} predictions look "smoothed". Essentially we found that while \unetmultistep{} had reasonably low MSE, it failed to capture high frequency detail and suffered from spectral bias. Thus, \unet{} was chosen as our surrogate model in \ourmethodshort{} across all experiments because it showed comparable cumulative MSE to \unetmultistep{} (Table \ref{tab:ns}).

\subsection{\ourmethodshort{} versus \unet{} Prediction}

\begin{figure}[h!]
\begin{center}
\includegraphics[width=\linewidth]{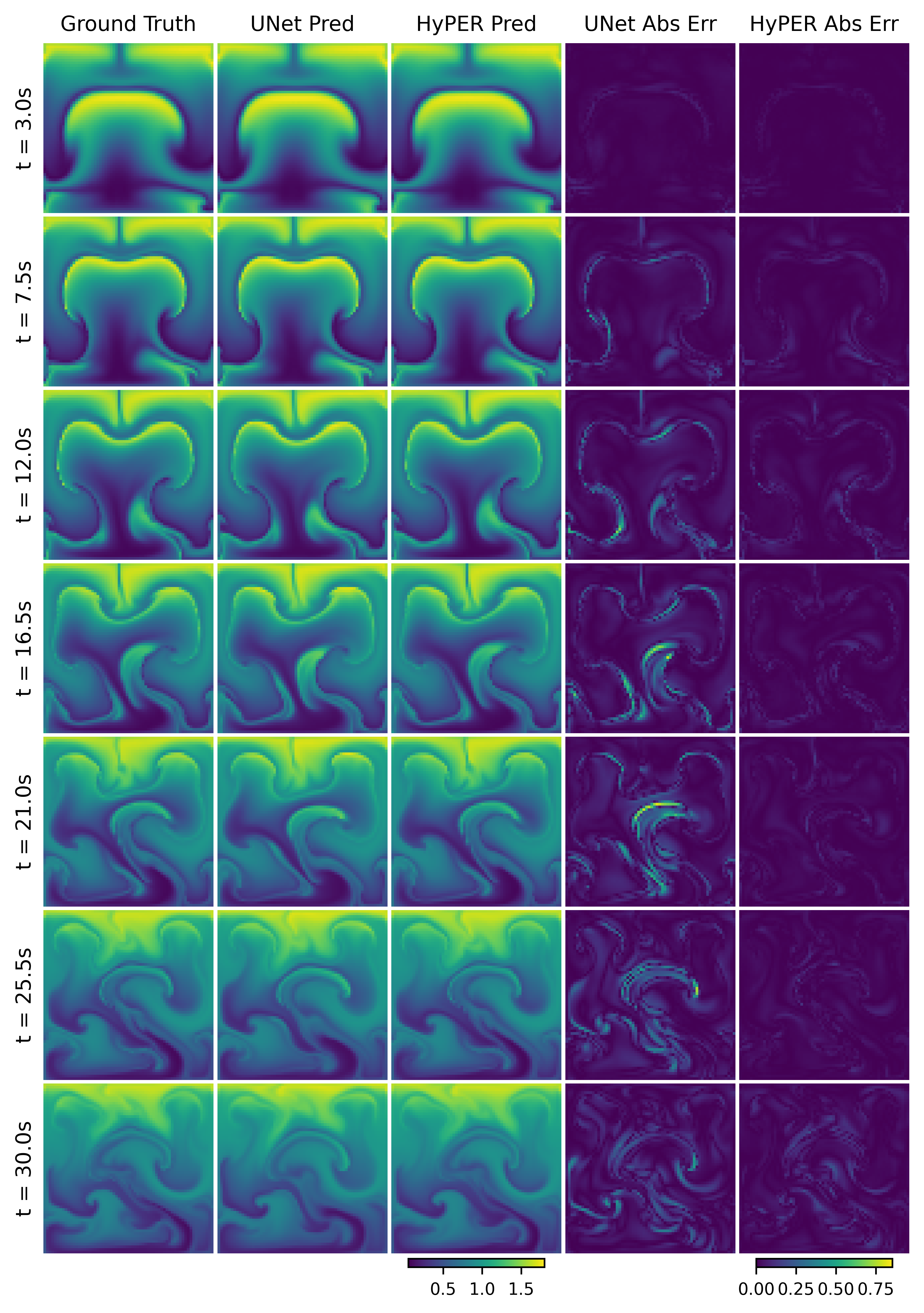}
\end{center}
\caption{Sample trajectory predictions and absolute errors of \ourmethodshort{} vs \unet{}. At various timesteps including $t = 21.0s$, we see that \unet{} has high error on the edges of the plume where high frequency behavior is present while \ourmethodshort{} successfully reduces this error.}
\label{fig:hyper-vs-unet}
\end{figure}

We contrast the predictions and errors of a \ourmethodshort{} rollout and \unet{} rollout in Figure \ref{fig:hyper-vs-unet}. Notice that at $t=\{16.5s, 21.0s\}$, \unet{} suffers from spectral bias and has high error on the plume edges where high frequency detail is required. We see that \ourmethodshort{} successfully mitigates this spectral bias error by incorporating the simulator, thereby significantly reducing rollout (cumulative) error over the full trajectory.

\subsection{Model Architectures and Parameters}
\label{app:surr-models}

\subsubsection{Model Training Details}

We build and train all our neural surrogate models using the PyTorch library on a single Nvidia RTX A6000 GPU. The \href{https://github.com/tum-pbs/PhiFlow}{$\Phi$Flow} simulator used in the Navier-Stokes experiment runs on the same GPU. The \href{https://github.com/OrchardLANL/DPFEHM.jl}{DPFEHM} Julia simulator runs using multi-threading on an Intel(R) Xeon(R) Platinum 8358 CPU @ 2.60GHz.

The \unet{}, \fno{}, \unetmultistep{}, \mpp{}, and \refiner{} baselines are trained with same number of samples as \ourmethodshort{} for the sake of fair comparison.
The \unetonly{} and \fnoonly{} baseline models are trained with a smaller dataset of 400 trajectories which are separate from the 400 trajectories \ourmethodshort{}'s RL policy is trained with.
All baselines are trained for 200 epochs.

We only train the HyPER RL ResNet model for 30 epochs to demonstrate that \ourmethodshort{} does not require extensive training to outperform the baselines.

\begin{table}[h]
\caption{Model training times and datasets for \surrogateonlyshort{} methods and \ourmethodshort{}.}
\label{tab:training-times}
\begin{center}
  \begin{tabular}{l|ccc}
    \toprule
    \textbf{Model} & \textbf{Training Time (Hours)} & \textbf{Epochs} & \textbf{Dataset Size (\# Trajectories)} \\
      \midrule
    \unetonly{} & 1.46 & 200 & 400 \\
    \fnoonly{} & 0.55 & 200 & 400 \\
    \unet{} & 3.29 & 200 & 800 \\
    \fno{} & 2.02 & 200.& 800 \\
    \unetmultistep{} & 4.10 & 200 & 800 \\
    \mpp{} & 9.58 & 200 & 800 \\
    \refiner{} & 3.66 & 200 & 800 \\
    \ourmethodshort{} & 5.21 & 30 & 400 \\
    \bottomrule
  \end{tabular}
\end{center}
\end{table}

\subsubsection{Model Parameters}

Our UNet model is built on top of the \href{https://github.com/pdearena/pdearena}{PDEArena} modern UNet architecture with wide residual blocks and training parameters in Table \ref{tab:unet-params}. 

\begin{table}[h]
\caption{UNet parameters.}
\label{tab:unet-params}
\begin{center}
  \begin{tabular}{l|c}
    \toprule
    \multirow{1}{*}{\textbf{Parameter}} &
      \multicolumn{1}{c}{\textbf{Value}} \\
      \midrule
    Model Size (\# parameters) & 12,295,233 \\
    Hidden Channels & 64 \\
    Activation Function & GELU \\
    Channel Multipliers & [1, 2, 2] \\
    Num Residual Blocks Per Channel & 2 \\
    Sinusoidal Time Embedding & Yes \\
    Learning Rate & 1e-4 \\
    Optimizer & Adam \\
    Loss Function & MSE \\
    Epochs & 200 \\
    \bottomrule
  \end{tabular}
\end{center}
\end{table}

Our FNO model is built using the \href{https://neuraloperator.github.io/dev/index.html}{neuraloperator} library and is constructed to have a similar number of parameters as our UNet for a fair comparison. See Table \ref{tab:fno-params} for details.

\begin{table}[h]
\caption{FNO parameters.}
\label{tab:fno-params}
\begin{center}
  \begin{tabular}{l|c}
    \toprule
    \multirow{1}{*}{\textbf{Parameter}} &
      \multicolumn{1}{c}{\textbf{Value}} \\
      \midrule
    Model Size (\# parameters) & 12,437,057 \\
    Number of Fourier Modes & 27 \\
    Hidden Channels & 64 \\
    Lifting Channels & 256 \\
    Projection Channels & 256 \\
    Learning Rate & 1e-5 \\
    Optimizer & Adam \\
    Loss Function & MSE \\
    Epochs & 200 \\
    \bottomrule
  \end{tabular}
\end{center}
\end{table}

The \ourmethodshort{} RL model is a lightly modified version of the ResNet34 model from the \href{https://github.com/pytorch/vision}{Torchvision} library. See Table \ref{tab:rl-params} for parameters.

\begin{table}[h]
\caption{RL ResNet parameters.}
\label{tab:rl-params}
\begin{center}
  \begin{tabular}{l|c}
    \toprule
    \multirow{1}{*}{\textbf{Parameter}} &
      \multicolumn{1}{c}{\textbf{Value}} \\
      \midrule
    Model Size (\# parameters) & 11,751,300 \\
    Number of Layers & [3, 4, 6, 3] \\
    Hidden Channels & 64 \\
    Sinusoidal. Time Embedding & Yes \\
    Activation Function & ReLU \\
    Learning Rate & 1e-5 \\
    Optimizer & Adam \\
    Reward Function & Equation \ref{eq:total-reward} \\
    Epochs & 30 \\
    \bottomrule
  \end{tabular}
\end{center}
\end{table}

The Multiple Physics Pretrained model (\mpp) is adapted from \href{https://github.com/PolymathicAI/multiple_physics_pretraining}{MPP} github. Note that we train this model for 200 epochs while \unet{} and \fno{} are only trained for 50. See Table \ref{tab:mpp-params} for details.

\begin{table}[h]
\caption{Multiple Physics Pretraining parameters.}
\label{tab:mpp-params}
\begin{center}
  \begin{tabular}{l|c}
    \toprule
    \multirow{1}{*}{\textbf{Parameter}} &
      \multicolumn{1}{c}{\textbf{Value}} \\
      \midrule
    Model Size (\# parameters) & 28,979,436 \\
    Patch Size & 16x16 \\
    Embedding Dimension & 384 \\
    Number of Axial Attention Heads & 6 \\
    Number of Transformer Blocks & 12 \\
    Epochs & 200 \\
    \bottomrule
  \end{tabular}
\end{center}
\end{table}

The PDE-Refiner model was adapted from \href{https://phlippe.github.io/PDERefiner/}{PDE-Refiner} to work with our UNet model. Note that we train this model for 200 epochs while \unet{} and \fno{} are only trained for 50. See Table \ref{tab:refiner-params} for details.

\begin{table}[h!]
\caption{PDE-Refiner parameters.}
\label{tab:refiner-params}
\begin{center}
  \begin{tabular}{l|c}
    \toprule
    \multirow{1}{*}{\textbf{Parameter}} &
      \multicolumn{1}{c}{\textbf{Value}} \\
      \midrule
    Model Size (\# parameters) & 12,378,241 \\
    Number of Refinement/Denoising Steps & 3 \\
    Minimum Noise Scale & 4e-7 \\
    Hidden Channels & 64 \\
    Activation Function & GELU \\
    Channel Multipliers & [1, 2, 2] \\
    Num Residual Blocks Per Channel & 2 \\
    Sinusoidal Time Embedding & Yes \\
    Learning Rate & 1e-4 \\
    Optimizer & Adam \\
    Loss Function & MSE \\
    Epochs & 200 \\
    \bottomrule
  \end{tabular}
\end{center}
\end{table}

\end{document}